\documentclass[11pt,a4paper]{article}
\usepackage[hyperref]{acl2021}
\usepackage{times}
\usepackage{latexsym}
\usepackage{multirow}
\usepackage{array}
\usepackage{graphicx} 
\usepackage{amstext}

\usepackage{microtype}

\title{How Knowledge Graph and Attention Help? \\ A Quantitative Analysis into Bag-level Relation Extraction}

\author{\textbf{Zikun Hu\textsuperscript{1} , Yixin Cao\textsuperscript{2} , Lifu Huang\textsuperscript{3} , Tat-Seng Chua\textsuperscript{1} }\\
\textsuperscript{1}National University of Singapore\\ \textsuperscript{2}S-Lab, Nanyang Technological University\\
\textsuperscript{3}Computer Science Department, Virginia Tech\\
\texttt{zikunhu@nus.edu.sg, yixin.cao@ntu.edu.sg}\\
\texttt{lifuh@vt.edu, chuats@comp.nus.edu.sg}}

\date{}

\begin{document}
\maketitle

\begin{abstract}

Knowledge Graph (KG) and attention mechanism have been demonstrated effective in introducing and selecting useful information for weakly supervised methods. However, only qualitative analysis and ablation study are provided as evidence. In this paper, we contribute a dataset and propose a paradigm to quantitatively evaluate the effect of attention and KG on bag-level relation extraction (RE). We find that (1) higher attention accuracy may lead to worse performance as it may harm the model's ability to extract entity mention features; (2) the performance of attention is largely influenced by various noise distribution patterns, which is closely related to real-world datasets; (3) KG-enhanced attention indeed improves RE performance, while not through enhanced attention but by incorporating entity prior; and (4) attention mechanism may exacerbate the issue of insufficient training data. Based on these findings, we show that a straightforward variant of RE model can achieve significant improvements (6\% AUC on average) on two real-world datasets as compared with three state-of-the-art baselines. Our codes and datasets are available at https://github.com/zig-kwin-hu/how-KG-ATT-help.
\end{abstract}

\section{Introduction}
\label{sec:introduction}


Relation Extraction (RE) is crucial for Knowledge Graph (KG) construction and population. Most recent efforts rely on neural networks to learn efficient features from large-scale annotated data, thus correctly extract the relationship between entities. To save the manual annotation cost and alleviate the issue of data scarcity, distant supervision relation extraction (DSRE) \cite{mintz2009distant} is proposed and becomes increasingly popular as it can automatically generate large-scale labeled data. DSRE is based on a simple yet effective principle: if there is a relation between two entities in KG, then all sentences containing mentions of both entities are assumed to express this relation and will form a sentence bag as its annotations.

\begin{figure}[htb]
  \centerline{\includegraphics[width=0.48\textwidth]{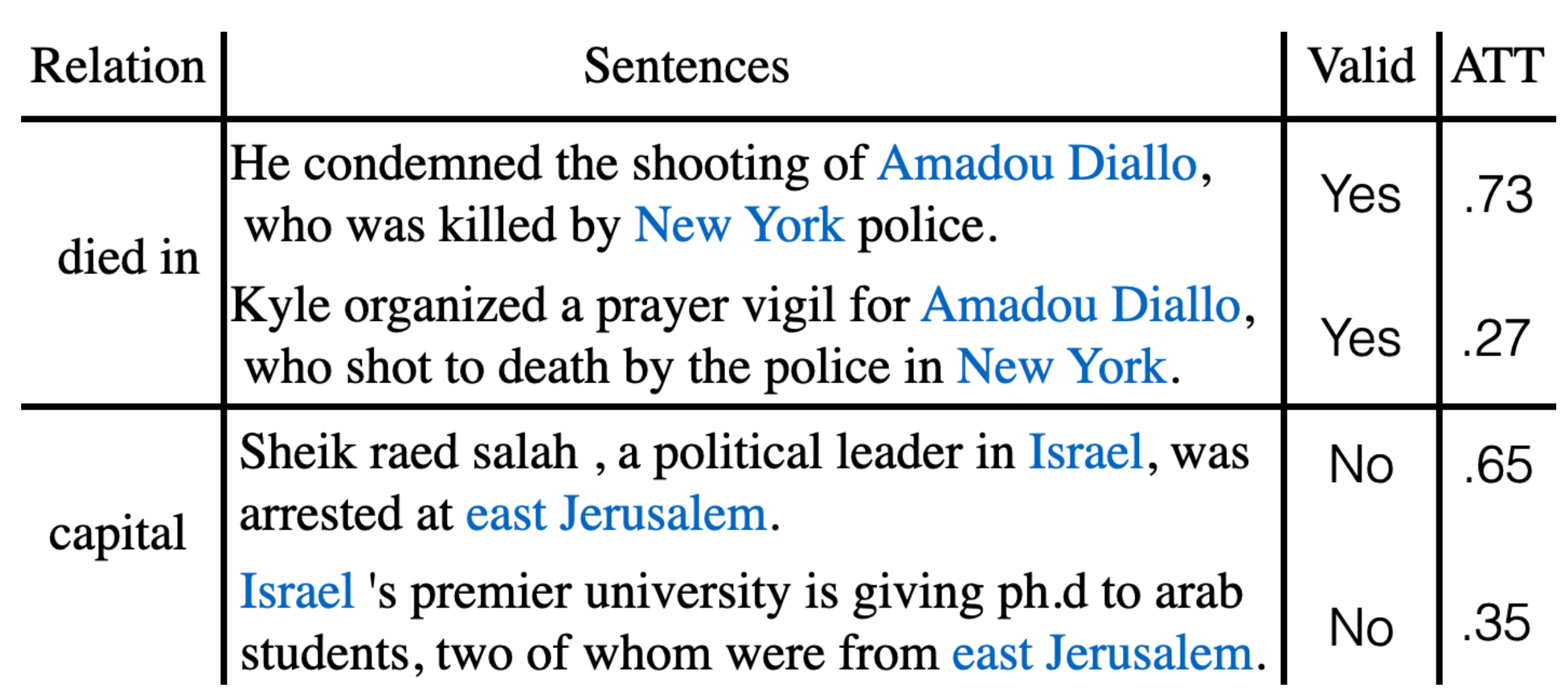}}

  \caption{Examples of disturbing bags in NYT-FB60K.}
  \label{fig:cases}
 \vspace{-0.2cm}
\end{figure}

Although effective, distant supervision may introduce noise to a sentence bag when the assumption fails --- some sentences are not describing the target relation \cite{zeng2015distant} (a.k.a. noisy annotation). To alleviate the negative impacts of noise, recent studies~\cite{lin2016neural,ji2017distant,du2018multi,li2020self} leveraged attention to select informative instances from a bag. Furthermore, researchers introduced KG embeddings to enhance the attention mechanism~\cite{hu2019improving,han2018neural}. The basic idea is to utilize entity embeddings as the query to compute attention scores, so that
the sentences with high attention weights are more likely to be valid annotations~\cite{zhang2019long}.
Previous studies have shown performance gain on DSRE with attention module and KG embeddings, however, it's still not clear how these mechanisms work, and, are there any limitations to apply them? 

In this paper, we aim to provide a thorough and quantitative analysis about the impact of both attention mechanism and KG on DSRE. By analyzing several public benchmarks including NYT-FB60K \cite{han2018neural}, we observe lots of \textbf{disturbing bags} --- all of the bag's sentences are valid or noisy annotations, which shall lead to the failure of attention. As shown in Figure-\ref{fig:cases}, all of annotations in the first disturbing bag are valid, while the learned attentions assign the second annotation with a very low weight, which suggests an inefficient utilization of annotations and exacerbates the data sparsity issue. Or, in the second bag, all sentences are noisy, can attention and KG still improve the performance? If so, how do they work and to what extent can they tolerate these disturbing bags? Answering these questions is crucial since this type of noise is common in practice. The unveiling of their working mechanism shall shed light on future research direction, not limited to DSRE.

To achieve this, we propose a paradigm based on newly curated DSRE benchmark, BagRel-Wiki73K extracted from FewRel~\cite{han2018fewrel} and Wikidata~\footnote{dumps.wikimedia.org/wikidatawiki/entities/20201109/}, for quantitative analysis of attention and KG. With extensive experiments, we conclude the following innovative and inspiring findings:

(1) The accuracy of attention is inversely proportional to the total noise ratio and disturbing bag ratio of training data; (2) attention effectively selects valid annotations by comparing their contexts with the semantics of relations, thus tends to rely more on the context to make predictions. However, it somehow lowers the model's robustness to noisy sentences that do not express the relation; (3) KG-enhanced attention indeed improves RE performance, surprisingly not via enhanced attention accuracy, but by incorporating entity features to reduce the demand of contexts when facing noise; (4) attention could hurt the performance especially when there is no sufficient training data.

Based on the above observations, we propose a new straightforward yet effective model based on pre-trained \textbf{B}ERT~\cite{devlin2018bert} for \textbf{RE} with \textbf{C}oncatenated KG \textbf{E}mbedding, namely \textbf{BRE+CE}.
Instead of in-bag attention, it breaks the bag and ensembles the results of all sentences belonging to the bag. For each sentence, we directly incorporate entity embeddings into BERT, rather than to enhance attentions, to improve the robustness of extracting both context and mention features.
\textbf{BRE+CE} significantly outperforms existing state-of-the-arts on two publicly available datasets, NYT-FB60K \cite{han2018neural} and GIDS-FB8K \cite{jat2018improving}, by 6\% AUC on average. We summarize our contributions as follows:

\begin{itemize}
    \item To the best of our knowledge, our proposed framework is the first work to quantitatively analyze the working mechanism of Knowledge Graph and attention for bag-level RE.
    \item We have conducted extensive experiments to inspire and support us with the above findings.
    \item We demonstrate that a straightforward method based on the findings can achieve improvements on public datasets.
\end{itemize}


\section{Related Work}
To address the issue of insufficient annotations, \citet{mintz2009distant} proposed distant supervision to generate training data automatically, which also introduces much noise. From then, DSRE becomes a standard solution that relies on multi-instance learning from a bag of sentences instead of a single sentence~\cite{riedel2010modeling,hoffmann2011knowledge}. Attention mechanism \cite{lin2016neural} accelerates this trend via strong ability in handling noisy instances within a bag~\cite{liu2017soft,du2018multi}. Aside from intra-bag attention, \newcite{ye2019distant} also designed inter-bag attention simultaneously handling bags with the same relation. To deal with only-one-instance bags, \newcite{li2020self} utilized a new selective gate (SeG) framework to independently assign weights to each sentence. External KG is also incorporated to enhance the attention module~\cite{han2018neural,hu2019improving}. However, due to the lack of sentence-level ground truth, it is difficult to quantitatively evaluate the performance of the attention module. Previous researchers tend to provide examples as case study.\footnote{\citet{shahbazi2020relation} claim to annotate each positive bag in NYT-FB60K, but haven't published their code and dataset.} Therefore, we aim to fill in this research gap by constructing a dataset and providing a framework for thorough analysis.

\section{Preliminary}
\label{sec:preliminary}
\textbf{Knowledge Graph} (KG) is a directed graph $G = \{E, R, T\}$, where E denotes the set of entities, $R$ denotes the set of relation types in $G$, and $T = \{(h, r, t)\}\subseteq E\times R \times E$ denotes the set of triples. KG embedding models, e.g., RotatE \cite{sun2019rotate}, can preserve the structure information in the learned vectors $\mathbf{e}_h$, $\mathbf{e}_t$ and $\mathbf{e}_r$. We adopt TransE \cite{bordes2013translating} in experiments.

\textbf{Bag-level relation extraction} (RE) takes a bag of sentences $B = \{s_1,s_2,\ldots, s_m\}$ as input. Each sentence $s_i$ in the bag contains the same entity pair $(h,t)$, where $h,t \in E$. The goal is to predict a relation $y \in R$ between $(h, t)$.

\textbf{Attention-based Bag-level RE} uses attention to assign a weight to each sentence within a bag. 
Given a bag $B$ from the dataset $D$, an encoder is first used to encode all sentences from $B$ into vectors $\{\mathbf{s'}_1,\mathbf{s'}_2,\ldots,\mathbf{s'}_m\}$ separately. Then, an attention module computes an attention weight $\alpha_i$ for each sentence and outputs the weighted sum of $\{\mathbf{s'}_i\}$ as $\mathbf{\overline{s}}$ to denote $B$:
\begin{equation}
    \omega_i = \mathbf{v}_y \cdot \mathbf{s'}_i
\end{equation}
\begin{equation}
    \alpha_i = \frac{\exp(\omega_i)}{\sum\limits_{j=1}^m \exp(\omega_j)}
\end{equation}
\begin{equation}
    \mathbf{\overline{s}} = \sum\limits_{i=1}^m \alpha_i\mathbf{s'}_i
\end{equation}
where $\mathbf{v}_y$ is the label embedding of relation $y$ in the classification layer, we denote this attention module as \textbf{ATT} in the rest of paper.

KG-enhanced attention aims to improve $\mathbf{v}_y$ with entities $\mathbf{e}_h$ and $\mathbf{e}_t$ \cite{han2018neural}:
\begin{equation}
    \mathbf{r}_{ht} = \mathbf{e}_h - \mathbf{e}_t
\end{equation}
\begin{equation}
    \omega_i = \mathbf{r}_{ht} \cdot \tanh(\mathbf{W}_s \mathbf{s'}_i + \mathbf{b}_s)
\end{equation}
where $\mathbf{r}_{ht}$ is regarded as latent relation embedding. We mark this way of computing $\omega_i$ as \textbf{KA}. $\mathbf{W}_s$ and $\mathbf{b}_s$ are learnable parameters.

Given a bag representation $\mathbf{\overline{s}}$, the classification layer further predicts a confidence of each relation:
\begin{equation}
    \overline{\mathbf{o}} = \mathbf{W}_b \overline{\mathbf{s}} + \mathbf{b}_b
\end{equation}
\begin{equation}
    P(y|B) = \text{Softmax}(\overline{\mathbf{o}})
\end{equation}
where $\overline{\mathbf{o}}$ is a logit vector. $\mathbf{W}_b$ and $\mathbf{b}_b$ are learnable parameters.
During training, the loss is computed by:
\begin{equation}
    L = - \sum\limits_{i=0}^n\text{log}(P(y_i|B_i))
    \label{eqlossbag}
\end{equation}
where $n$ is the number of training bags in $D$. Since the classification layer is linear, we can rewrite the bag's logit vector $\overline{\mathbf{o}}$ using a weighted sum of each sentence's logit vector $\mathbf{o}$:
\begin{equation}
    \mathbf{o}_i = \mathbf{W}_b \mathbf{s'}_i + \mathbf{b}_b
\end{equation}
\begin{equation}
    \overline{\mathbf{o}} = \sum\limits_{i=1}^m \alpha_i\mathbf{o}_i
    \label{eqreview}
\end{equation}

From equation \ref{eqreview}, we can see that the model's output on the whole bag depends on three aspects: (1) the model's output on valid sentences within the bag; (2) the model's output on noisy sentences within the bag; (3) the attention weight assigned to valid sentences and noisy ones.

\section{Benchmark}
To quantitatively evaluate the effect of attention and KG on Bag-level RE, we first define two metrics to measure the noise pattern (Section \ref{sec:noisemetrics}). Then, we construct a KG and a Bag-level RE dataset (Section \ref{sec:datasetconstruct}). Finally, we introduce a general evaluation framework to assess attention, KG and the entire RE model (Section \ref{evalframework}).

\subsection{Metrics Describing Noise Pattern}
\label{sec:noisemetrics}
To analyze how attention module functions on different noise patterns, we first design 2 metrics to describe the noise pattern: \textbf{Noise Ratio} (NR) and \textbf{Disturbing Bag Ratio} (DR).

\paragraph{Noise Ratio (NR)} represents the proportion of noisy sentences in the dataset. Given a bag $B_i$ and its relation label $y_i$, a sentence $s_{ij}\in B_i$ is noisy if its context does not express $y_i$. Suppose $\text{Isn}(s_{ij}, y_i)$ is an indicator function to tell whether $s_{ij}$ is noise. Then \textbf{NR} is defined as:
\begin{equation}
    \text{NR} = \frac{\sum\limits_{i=1}^n\sum\limits_{j=1}^{|B_i|}\text{Isn}(s_{ij},y_i)}{\sum\limits_{i=1}^n|B_i| }
\end{equation}
where $|B_i|$ is the size of $B_i$, $n$ is the total number of bags.

\paragraph{Disturbing Bag Ratio (DR)} means the proportion of disturbing bags in the dataset. A bag is disturbing if all sentences in it are valid or all sentences are noisy. Formally, we use function $\text{Isd}(B_i)$ to indicate whether a bag is disturbing or not:
\begin{equation}
    \text{Isd}(B_i) = \prod_{j=1}^{|B_i|} \text{Isn}(s_{ij},y_i)+\prod_{j=1}^{|B_i|} (1-\text{Isn}(s_{ij},y_i))
\end{equation}
Then we define \textbf{DR} as follows:
\begin{equation}
    \text{DR} = \frac{\sum\limits_{i=1}^n\text{Isd}(B_i)}{n}
\end{equation}

\begin{figure*}[htb]
  \centerline{\includegraphics[width=0.96\textwidth]{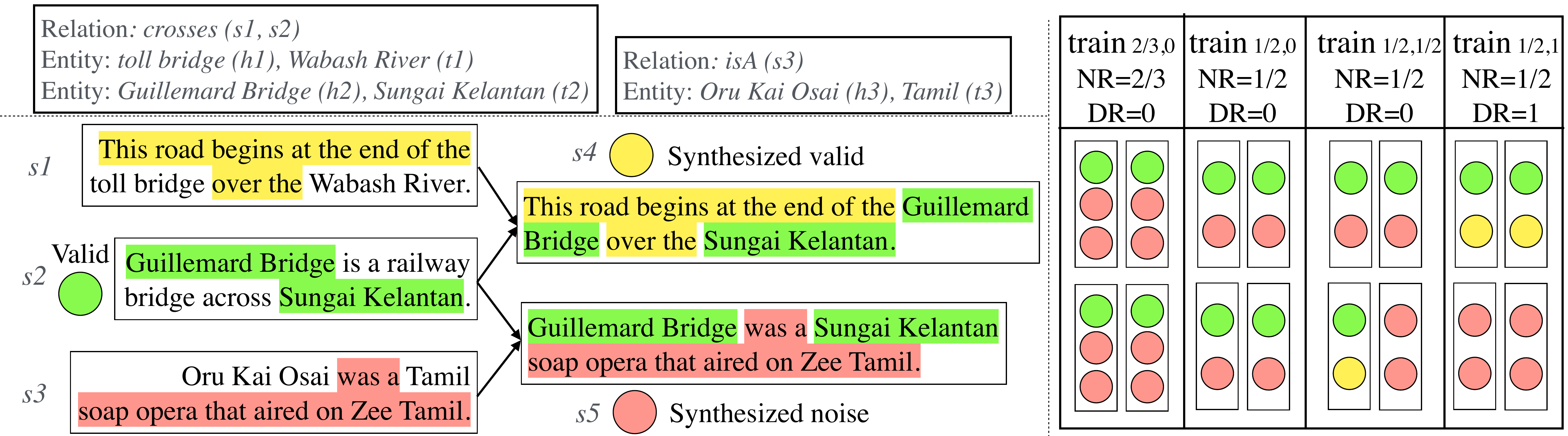}}

  \caption{Left: Process of synthesizing the valid sentence with correct context and the noisy sentence with wrong context. Right: Visualization of different train sets of different noise patterns, the four sets from left to right are named as $\text{train}_{\frac{2}{3},0}$,$\text{train}_{\frac{1}{2},0}$,$\text{train}_{\frac{1}{2},\frac{1}{2}}$ and $\text{train}_{\frac{1}{2},1}$.}
  \label{fig:dataset}
\end{figure*}

\subsection{Dataset Construction}
\label{sec:datasetconstruct}
Based on FewRel and Wikidata, we construct a Bag-level RE dataset containing multiple training sets with different noise patterns, a test set and a development set. For each sentence in the bags, there is a ground truth \textbf{attention label} indicating whether it is a valid sentence or noise. We also construct a KG containing all entities in the RE dataset by retrieving one-hop triples from Wikidata.

\paragraph{Synthesize Sentence} FewRel is a sentence-level RE dataset, including 80 relations. For each relation, there are 700 valid sentences. Each sentence has a unique entity pair. Every sentence along with its entities and relation label form a tuple $(s, h, t, y)$. We thus synthesize valid and noisy sentences for the same entity pair for data augmentation.

The first step is to divide sentences of each relation into 3 sets: $\text{train}_{\text{FewRel}}$, $\text{test}_{\text{FewRel}}$ and $\text{dev}_{\text{FewRel}}$, where each set has 500, 100 and 100 sentences. Then, for each tuple $(s, h, t, y)$ in the set, we aim to augment it to a bag $B$, where all of its sentences contain $(h, t)$. Besides, the sentences in $B$ are either the original $s$, or a synthesized valid sentence, or a synthesized noisy sentence. We synthesize sentences in the form of $(s', h, t, y, z)$, where $z$ denotes the attention label (1 for valid, 0 for noisy). In specific, to synthesize a sentence, we randomly replace the source pair of entity mentions with other target entity pairs while keeping the context unchanged. Thus, if the contexts express the same relation type with the entity pair, we can automatically assign an attention label. 

We illustrate the synthesizing process in Figure~\ref{fig:dataset}. $(s_2, h_2, t_2, \textit{crosses})$ is a sentence from $\text{train}_{\text{FewRel}}$.  To generate a valid sentence, we randomly select another sentence $(s_1, h_1, t_1, \textit{crosses})$ which is labeled with the same relation as $s_2$ from $\text{train}_{\text{FewRel}}$. Then we replace its entity mentions $h_1$ and $t_1$ as $h_2$ and $t_2$. The output is $(s_4, h_2, t_2, \textit{crosses}, 1)$. Since its context correctly describe \textit{crosses}, we regard $s_4$ as valid. For the noisy sentence, we randomly select a sentence $(s_3, h_3, t_3, \textit{isA})$ under another relation. With similar process for $s_4$, we get a synthesize sentence $(s_5, h_2, t_2, \textit{crosses}, 0)$. Because the context of $s_5$ does not express target relation, we label it as a noise.

\paragraph{Training Sets with Different Noise Patterns} As defined in Section~\ref{sec:noisemetrics}, we use NR and DR to measure the noise pattern of Bag-level RE dataset. By controlling the number of synthesized noisy sentences in each bag and the total ratio of noise among all sentences, we can construct several training sets with different patterns. In the following sections, we denote a training set of which the NR is $x$ and DR is $y$ as $\text{train}_{x,y}$. Higher $x$ and $y$ indicate noisy sentences and disturbing bags account for larger proportion. 

For example, in Figure~\ref{fig:dataset}, assuming there are 4 sentences in $\text{train}_{\text{FewRel}}$, for each sentence, we synthesize two noisy sentences that form the bag together with the original sentence. Thus each bag contains 3 sentences: 1 valid and 2 noisy, and its NR is 2/3 and DR is 0. For the other 3 sets, the number of synthesized noisy sentences equals the sum of original valid sentences and synthesized valid sentences. Thus they all have a NR of 1/2. Since we define bags containing no valid sentences or no noisy sentences as disturbing bags, the third set and fourth set have 2 and 4 disturbing bags, with a DR of 1/2 and 1, respectively.

\paragraph{Test Set and Development Set} We also construct a test and a development set. Similar as the second set in Figure~\ref{fig:dataset}, each bag in the test/dev sets contains two sentences, the NR of both sets is 1/2 while the DR is 0. I.e., in every bag of test/dev sets, there is one valid sentence and one noisy sentence. Instead of multiple test sets of different noise patterns, we only have one test set so that the evaluation of different models is consistent. To avoid information leak, when construct $\text{train}_{x,y}$, test and development sets, the context of synthesized sentences only come from $\text{train}_{\text{FewRel}}$, $\text{test}_{\text{FewRel}}$ and $\text{development}_{\text{FewRel}}$, respectively.

The final BagRel contains 9 train sets, 1 test and 1 development set, as listed in Table~\ref{tabble:bagrel}. The NR of the training sets has three options: 1/3, 1/2 or 2/3, and similarly, DR can be 0, 1/2 or 1. The NR of both test and development sets are 1/2, while their DR are 0. All data sets contain 80 relations. For training sets whose NR are 1/3, 1/2 and 2/3, every bag in these sets contains 3, 2 and 3 sentences, respectively. 
\begin{table}[htb]
\small
\centering
\begin{tabular}{cccc}
Dataset & \# Noisy Sentence& \# Sentence&\# Bag \\ \hline
$\text{train}_{\frac{1}{3},(0,\frac{1}{2}, 1)}$ &40K& 120K&40K \\ 
$\text{train}_{\frac{1}{2},(0,\frac{1}{2}, 1)}$ &40K& 80K&40K \\
$\text{train}_{\frac{2}{3},(0,\frac{1}{2}, 1)}$ &80K& 120K&40K \\ 
$\text{dev}_{\frac{1}{2},0}$ &8K& 16K&8K \\
$\text{test}_{\frac{1}{2},0}$ &8K& 16K&8K \\ 
\hline
\end{tabular}
\caption{Statistics of 11 sets of BagRel-Wiki73K, where $\text{train}_{c,(x, y, z)}$ denotes three sets of $\text{train}_{c,x}$, $\text{train}_{c,y}$, and $\text{train}_{c,z}$.}
\label{tabble:bagrel}
\vspace{-0.4cm}
\end{table}

\paragraph{KG Construction} To evaluate the impact of KG on attention mechanism, we also construct a KG based on Wikidata. Denoting the set of entities appearing in FewRel as $E$, we link each entity in $E$ to Wikidata by its Freebase ID, and then extract all triples $T = {(h, r, t)}$ in Wikidata where $h,t\in E$. To evaluate the effect of structural information from KG, we also construct a random KG whose triple set is $\hat{T}$. Specifically, for each triple $(h, r, t)$ in $T$, we corrupt it into $(h, \hat{r}, t)$ by replacing $r$ with a random relation $\hat{r} \neq r$. Thus the prior knowledge within the KG is destroyed. KG-73K and KG73K-random have the same scale: 72,954 entities, 552 relations and 407,821 triples.

Finally, we obtain BagRel-Wiki73K, including the Bag-level RE sets and KG-73K.

\subsection{Evaluation Framework}
\label{evalframework}
We first define several measurements to evaluate the effect of the attention mechanism and KG: \textbf{Attention Accuracy} (AAcc), \textbf{Area Under precision-recall Curve} (AUC),  \textbf{AUC on Valid sentences} (AUCV) and \textbf{AUC on Noisy sentences} (AUCN).

\paragraph{AAcc} measures the attention module's ability to assign higher weights to valid sentences than noisy sentences. Given a \textbf{non-disturbing bag} (a bag containing both valid and noisy sentences) $B_i = \{(s_j,h_i,t_i,y_i, z_j)\}$ and the predicted probability distribution $\mathbf{p_i}$, the AAcc of this bag is calculated by the following formula: 
\begin{equation}
        \text{AAcc}_i = \frac{\sum\limits_{j=1}^m\sum\limits_{k=1}^m\textbf{I}(z_j)\textbf{I}(1-z_k)\textbf{I}(p_{ij}>p_{ik})}{\sum\limits_{j=1}^m\textbf{I}(z_j)\sum\limits_{j=1}^m\textbf{I}(1-z_j)}
\end{equation}
where $m=|B_i|$ is the size of $B_i$, \textbf{I}($\cdot$) is an indicator function which returns 1 or 0 if the input is True or False. By $\sum\limits_{j=1}^m\textbf{I}(z_j)\sum\limits_{j=1}^m\textbf{I}(1-z_j)$, we count how many valid-noisy sentence pairs contained in $B_i$. With $\sum\limits_{j=1}^m\sum\limits_{k=1}^m\textbf{I}(z_j)\textbf{I}(1-z_k)\textbf{I}(p_{ij}>p_{ik})$, we count how many pairs show higher weight on the valid sentence. Then the AAcc of the whole data set is computed as
$\text{AAcc} =(\sum\limits_{i=1}^n\text{AAcc}_i)/n$ where n is the number of bags in the data set.

AAcc is designed specifically for non-disturbing bags. On disturbing bags, with all sentences noisy or valid, it is meaningless to evaluate attention module's performance. So in test/dev sets of our BagRel-Wiki73k, all bags are non-disturbing bags. Then without distraction, the evaluation results can better present how the attention module works.

\paragraph{AUC} is a standard metric to evaluate DSRE model's performance on bag-level test set. As mentioned in section~\ref{sec:preliminary}, attention-based model's performance on non-disturbing bags relies on three aspects: (1)AAcc, (2) model's performance on valid sentences and (3) model's performance on noisy sentences. So we use \textbf{AUCV} and \textbf{AUCN} to measure the second and the third aspects, respectively. The difference between AUC and AUCV is that AUC is computed on the original test set $D = \{B_i\}$, while AUCV is \textbf{AUC} computed on the \textbf{V}alid-only test set $D^v = \{B_i^v\}$. Compared with $B_i$, $B_i^v$ has the same label but removes all noisy sentences within it. Thus there is no noisy context feature in $D^v$, then models can utilize both entity mentions and contexts to achieve a high AUCV.  On the opposite, AUCN is \textbf{AUC} computed on the \textbf{N}oise-only test set $D^n = \{B_i^n\}$, where $B_i^n$ removes all valid sentences in $B_i$. Since all context features in $D^n$ are noisy, to achieve a high AUCN, models have to ignore context and rely more on mention features to make predictions.

AUC, AUCV and AUCN range from $0$ to $1$, and a higher value of the 3 metrics indicates that a model makes better prediction on the whole bag, valid sentences and noisy sentences, respectively.

\section{Method}
\label{sec:method}
To evaluate the effects of attention and KG, we design two straightforward Bag-level RE models without the attention module, $\textbf{BRE}$ and $\textbf{BRE}$\textbf{+CE}. By comparing their performance with $\textbf{BRE}$\textbf{+ATT} (BRE with attention module) and $\textbf{BRE}$\textbf{+KA} (BRE with KG-enhanced attention module), we can have a better understanding of the roles of ATT and Knowledge-enhanced ATT.

$\text{BRE}$ uses BERT \cite{devlin2018bert} as the encoder. Specifically, we follow the way described in~\cite{peng2020learning,soares2019matching}: entity mentions in sentences are highlighted with special markers before and after mentions. Then the concatenation of head and tail entity representations are used as the representation $s'$. Since $\text{BRE}$ does not have attention mechanism, it breaks the bags and compute loss on each sentence:
\begin{equation}
    L = -\sum\limits_{i=1}^n\sum\limits_{j=1}^{|B_i|}\text{log}(P(y_i|s_{ij}))
\end{equation}
\begin{equation}
    P(y_i|s_{ij}) = \text{softmax}(\mathbf{W}_b s'_{ij} + \mathbf{b}_b)
\end{equation}
$\text{BRE}$ can be viewed as a special case of $\text{BRE}$+ATT. Its attention module assigns all sentences in all bags with the same attention weight 1. During inference, given a bag, $\text{BRE}$ uses the mean of each sentence's prediction as the whole bag's prediction:
\begin{equation}
     P(y_i|B_i) = (\sum\limits_{j=1}^{|B_i|} P(y_i|s_{ij}))/|B_i|
\end{equation}

$\text{BRE}$+CE concatenates an additional feature vector $\mathbf{r}_{ht}$ with BERT output, where $\mathbf{r}_{ht}$ is defined based on entity embeddings of $h$ and $t$. The concatenated vector is used as the representation of the sentence and fed into the classification layer.

\section{Experiment}
We apply our proposed framework on BagRel-Wiki73K and two real-world datasets to explore the following questions:

\begin{itemize}
\setlength{\itemindent}{-1em}
    \item How noise pattern affects the attention module?
    \item Whether attention mechanism promotes RE model's performance?
    \item How KG affects the attention mechanism?
    \item Whether attention aggravates data sparsity?
\end{itemize}


\subsection{Experimental Setup}
For fair comparison, all of baselines share the same encoding structure as $\text{BRE}$. The attention-based models include $\text{BRE}$+ATT,$\text{BRE}$+KA and $\text{BRE}$+SeG, where SeG \cite{li2020self} is an advanced attention mechanism which achieves the state-of-the-art performance on NYT-FB60K. Briefly, SeG uses sigmoid instead of softmax to compute attention weights of each instance in a bag.
The models without attention are $\text{BRE}$ and $\text{BRE}$+CE. To check the effect of noise pattern, we train model on different train sets. As a reminder, $\text{train}_{x,y}$ is a train set whose NR and DR is $x$ and $y$, respectively.


\subsection{Noise Pattern v.s. Attention Accuracy}
\begin{figure}[htb]
 
  \centerline{\includegraphics[width=0.48\textwidth]{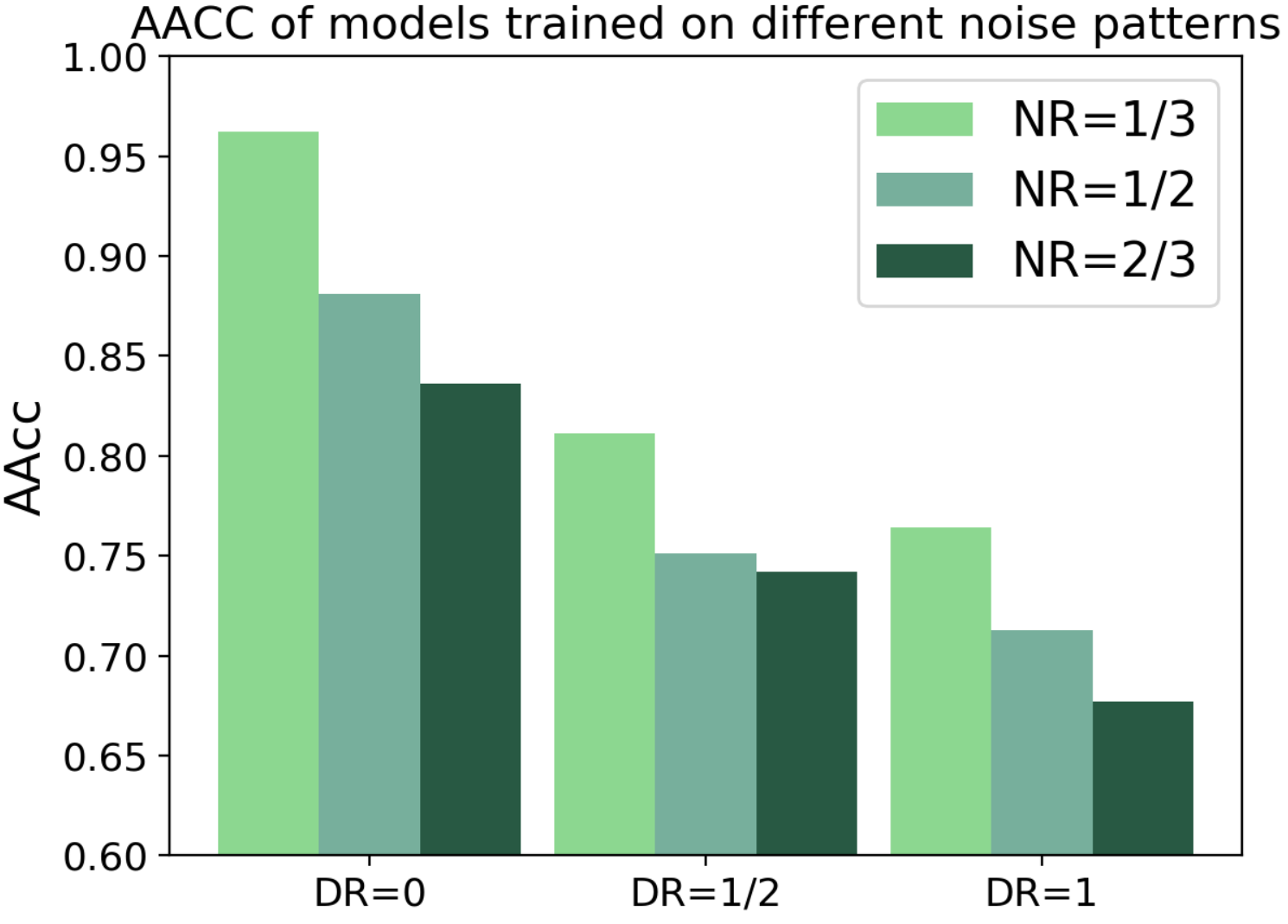}}

  \caption{Attention accuracy (AAcc) on the test set of  BagRel-Wiki73K. The results are collected with $\text{BRE}$+ATT trained on train sets of various noise patterns. The x axis denote train sets of different Disturbing bag Ratio (DR). The different colors indicate various Noise Ratio (NR).}
  \label{fig:attnacc}
 \vspace{0cm}
\end{figure}

We train $\text{BRE}$+ATT on 9 different training sets with different noise patterns. As shown in Figure \ref{fig:attnacc}, we can see that: (1) higher noise ratio (NR) makes the model harder to highlight valid sentences, leading to a lower attention accuracy (AAcc); (2) higher disturbing bag ratio (DR) results in lower AAcc, indicating that disturbing bags challenge the attention module. Based on these results, we claim that the noise pattern within the training set largely affects the attention module's effectiveness. 

\subsection{Attention v.s. RE Performance}

 \begin{table}[htbp]
 \small
\centering
\begin{tabular}{c|c|c|c|c}

\textbf{Model} & \textbf{AUC}& \textbf{AAcc}& \textbf{AUCV}&\textbf{AUCN} \\ \hline
$\text{BRE}$-$\text{train}_{\frac{1}{2},0}$  &.910& NA&.932&.850  \\
$\text{BRE}$+ATT-$\text{train}_{\frac{1}{2},0}$ &.878& .881&.941&.434 \\
$\text{BRE}$+ATT-$\text{train}_{\frac{1}{2},\frac{1}{2}}$ &.897&.751&.932&.711\\
$\text{BRE}$+ATT-$\text{train}_{\frac{1}{2},1}$ &.896&.713&.925&.759\\
\end{tabular}
\caption{Test results of models trained on different train set. In the Model column, X-Y means model X trained on train set Y. Among 3 train sets, $\text{train}_{\frac{1}{2},1}$ has the most disturbing bags, while $\text{train}_{\frac{1}{2},0}$ has no such bag.}  
\label{fewrel-1}
\vspace{-0.4cm}
\end{table}

To quantitatively analyze the effect of attention mechanism, we compare the performance of $\text{BRE}$ and $\text{BRE}$+ATT in Table \ref{fewrel-1}, keeping other variables of the model unchanged. Particularly, a higher AUCV indicates the stronger ability of the model itself --- in an ideal setting without any noise, and a higher AUCN indicates higher robustness of model to noise. Surprisingly, when using the same training set $\text{train}_{\frac{1}{2},0}$, the AUC of the attention-enhanced model is lower than the AUC of the model without attention ($0.878$ v.s. $0.910$). In addition, $\text{BRE}$+ATT has lowest AUC using $\text{train}_{\frac{1}{2},0}$, which has no disturbing bags. The highest AAcc ($0.881$) also suggests that the attention module does effectively select valid sentences. \textbf{Why the most effective attention module leads to the worst performance?} The reason is that $\text{BRE}$+ATT-$\text{train}_{\frac{1}{2},0}$ has a much lower AUCN, which indicates that it is less robust to noisy sentences.

\begin{figure}[tb]
 \vspace{-0.8cm}
  \centerline{\includegraphics[width=0.48\textwidth]{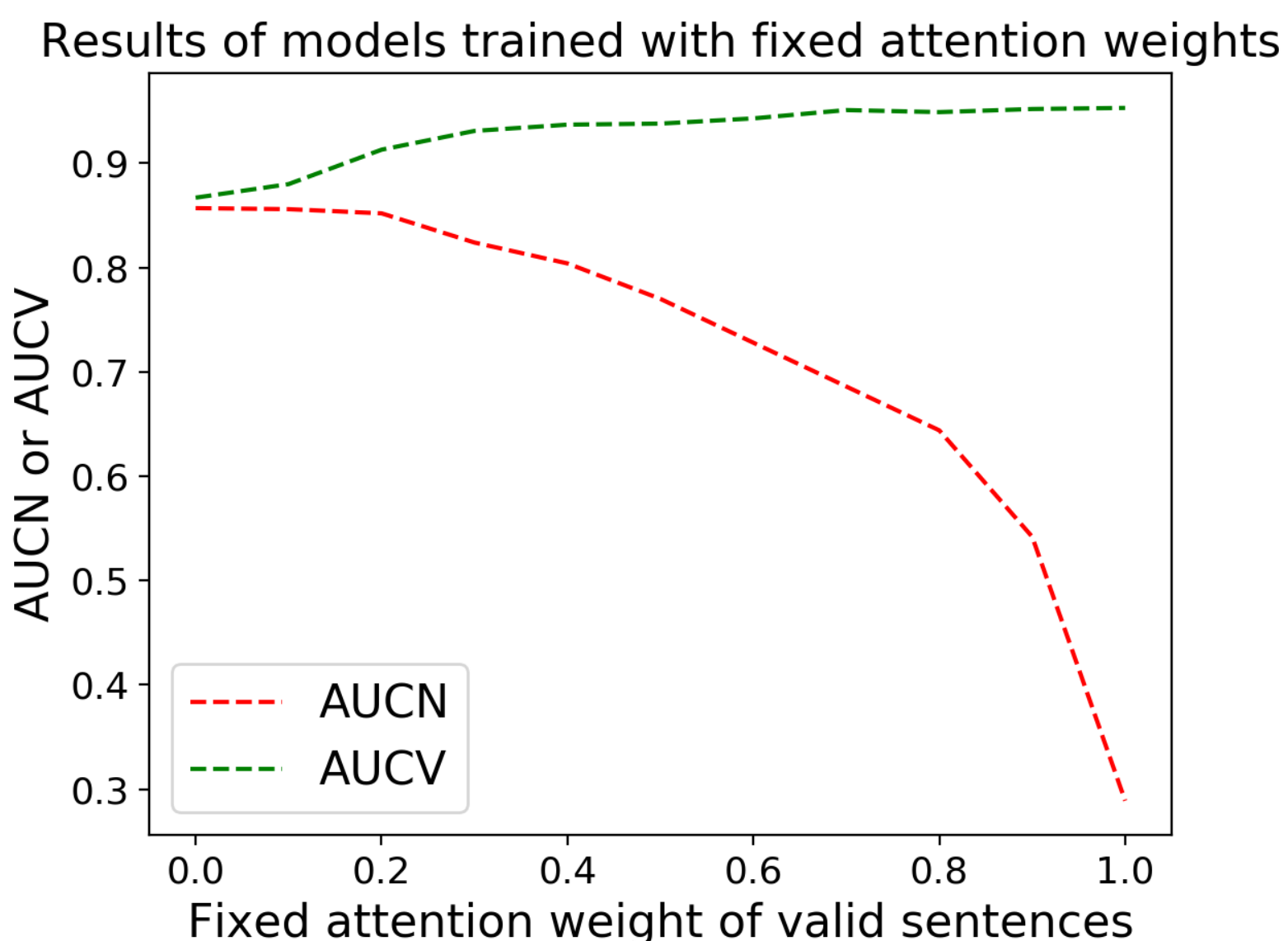}}

  \caption{AUCV and AUCN results of  $\text{BRE}$+ATT-$\text{train}_{\frac{1}{2},0}$ trained with fixed attention weights.}
  \label{fig:fixweight}
 \vspace{-0.6cm}
\end{figure}

Is it true that \textbf{an effective attention module shall hurt model's robustness to noise}? This is actually against our intuition.
To answer it, we draw Figure~\ref{fig:fixweight} by assigning fixed attention weights to sentences during training. Specifically, each bag in $\text{train}_{\frac{1}{2},0}$ has a valid sentence and a noisy sentence, and we assign fixed attention weight $\alpha$ to the valid and $1-\alpha$ to the noisy one, instead of computing $\alpha$ with attention module. Then we test the resulting model's AUCN and AUCV performance. We can see that when the valid sentences receive higher attention weights, the AUCV curve rises slightly, indicating the model's performance indeed gets enhanced. Meanwhile, the AUCN curve goes down sharply. This demonstrates the effective attention weakens the model's robustness to noise. The reason is that the model with a high-performance attention module prefers to utilize context information instead of entity mention features. Thus, it usually fails if most contexts are noisy.
Thus we can explain the results in Table \ref{fewrel-1}. $\text{train}_{\frac{1}{2},0}$ has the highest AAcc, indicating that it assigns very low weights to noisy sentences. Thus the gain from AUCV can not make up the loss from AUCN, resulting a worse AUC.

In conclusion, attention module can effectively select valid sentences during training and test. But it has an underlying drawback that it might hurt the model's ability to predict based on entity mention features, which are important in RE tasks \cite{li2020self} \cite{peng2020learning}, leading to worse overall performance.

 \subsection{KG v.s. Attention}
  \begin{table}[htbp]
 \small
\centering
\begin{tabular}{c|p{0.6cm}|p{0.6cm}|p{0.7cm}|p{0.7cm}}

\textbf{Model} & \textbf{AUC}& \textbf{AAcc}& \textbf{AUCV}&\textbf{AUCN} \\ \hline
$\text{BRE}$+ATT-$\text{train}_{\frac{1}{2},0}$ &.878& .881&.941&.434 \\
$\text{BRE}$+$\text{KA}_{\text{rand}}$-$\text{train}_{\frac{1}{2},0}$ &.915&.762&.936&.659\\
$\text{BRE}$+KA-$\text{train}_{\frac{1}{2},0}$  &.932& .857&.936&.560  \\
$\text{BRE}$+KA-$\text{train}_{\frac{1}{2},\frac{1}{2}}$ &.924&.720&.928&.723\\
$\text{BRE}$+KA-$\text{train}_{\frac{1}{2},1}$ & .913 &.617&.916&.761\\
\hline
\hline
$\text{BRE}$+CE-$\text{train}_{\frac{1}{2},0}$  &.915& NA&.935&.856  \\
$\text{BRE}$+CE-$\text{train}_{\frac{1}{2},\frac{1}{2}}$ &.919&NA&.939&.849   \\
$\text{BRE}$+CE-$\text{train}_{\frac{1}{2},1}$ &.918&NA&.941&.845   \\
\end{tabular}
\caption{Results of models trained on different train set. In the Model column, X-Y means model X trained on train set Y. $\text{BRE}$+$\text{KA}_{\text{rand}}$ uses entity embeddings learned on KG-73K-random for the attention module.}  
\label{fewrel-2}
\vspace{-0.4cm}
\end{table}
To measure KG's effect on the combined with attention mechanism, we compare the results of KA with ATT, while keeping other parts of the model unchanged. As shown in Table \ref{fewrel-2}. When trained on $\text{train}_{\frac{1}{2},0}$, the KG-enhanced model (KA-$\text{train}_{\frac{1}{2},0}$) has lower AAcc than the model without KG (ATT-$\text{train}_{\frac{1}{2},0}$) ($0.857$ v.s. $0.881$), while the AUC is higher ($0.932$ v.s. $0.878$). This is because the KA version has a higher AUCN ($0.560$) and comparable AUCV and AAcc. Thus, the KG-enhanced model achieves better performance on noisy bags, leading to a better RE performance.

In addition, comparing Table \ref{fewrel-1} and Table \ref{fewrel-2}, KA shows lower AAcc and higher AUCN than ATT on all three train sets. This also demonstrates that KG does not promote model's performance by improving attention module's accuracy, but by enhancing the encoder and classification layer's robustness to noisy sentences. This makes sense because the information from KG focuses on entities instead of contexts. By incorporating KG, the model relies more on entity mention features instead of noisy contexts feature, thus becomes better at classifying noisy sentences.

Moreover, comparing $\text{BRE}$+$\text{KA}_{\text{rand}}$'s performance with $\text{BRE}$+KA on $\text{train}_{\frac{1}{2},0}$, we can observe that after incorporating entity embeddings learned from a random KG, $\text{BRE}$+$\text{KA}_{\text{rand}}$ has a much lower attention accuracy. This indicates that misleading knowledge would hurt attention mechanism.
 
\subsection{Attention v.s. Data Sparsity}
\begin{figure}[htb]
 \vspace{-0.4cm}
  \centerline{\includegraphics[width=0.48\textwidth]{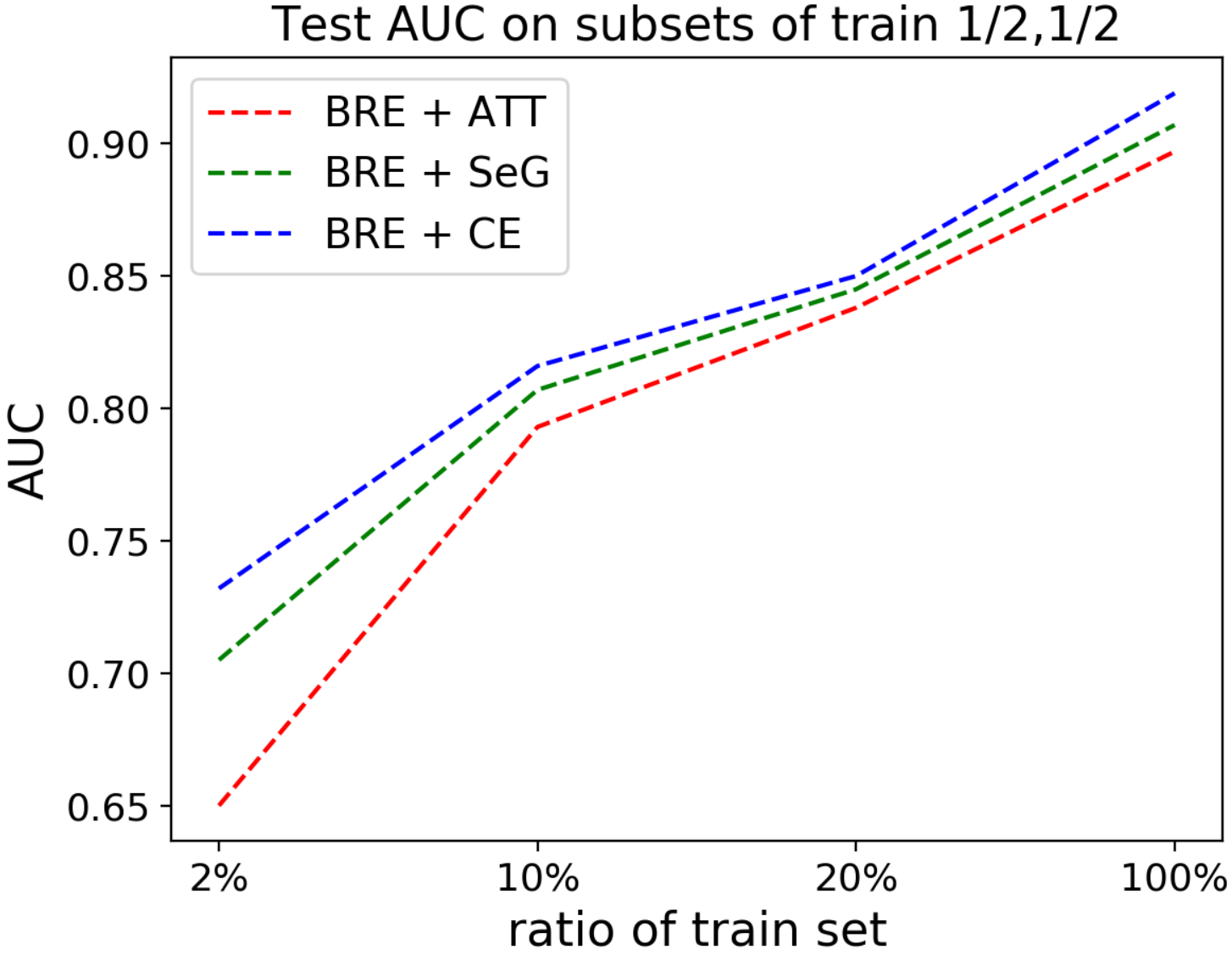}}

  \caption{AUC test results of models trained on 4 subsets of BagRel-Wiki73K's $\text{train}_{\frac{1}{2},\frac{1}{2}}$ set. The 4 subsets contain 2\%, 10\%, 20\% and 100\% bags of $\text{train}_{\frac{1}{2},\frac{1}{2}}$ set.}
  \label{fig:smalltrain}
 \vspace{-0.4cm}
\end{figure}

Attention module assigns low weights to part of training sentences. When training data is insufficient, not making full use of all training examples could aggravate the data sparsity issue. Thus we compare performance of models trained on subsets of $\text{train}_{\frac{1}{2},\frac{1}{2}}$. From Figure~\ref{fig:smalltrain}, we can see that along with the decreasing size of training data, the performance gap between $\text{BRE}$+ATT and $\text{BRE}$+CE becomes larger. This is because the latter one fully utilizes every example by assigning the same weight 1 to all sentences. We also check each model's attention weights. $\text{BRE}$+SeG assigns all sentences with weights $>0.9$, so its performance drop is similar to the model without attention. Thus, we claim that traditional attention mechanism could exacerbate the model's ability to insufficient data. This motivates us a better attention mechanism for few-shot settings. We leave it in the future.

 \subsection{Stability of Attention v.s. Noise Pattern}
From results in Table \ref{fewrel-1} and Table \ref{fewrel-2}, we can see that the performance of BRE+CE is stable when the ratio of disturbing bags changes. In comparison, BRE+ATT and BRE+KA show varying results across different train sets. On $\text{train}_{\frac{1}{2},1}$ which has the most disturbing bags, BRE+CE outperforms BRE+ATT and BRE+KA, demonstrating that BRE+CE could be a competitive method for Bag-level DSRE.
 
 \subsection{Results on Real-world Datasets}
 \begin{table}[htbp]
 \small
\centering
\begin{tabular}{c|c|c}

\textbf{Model} & \textbf{NYT-FB60K}& \textbf{GIDS-FB8K} \\ \hline
JointE &.408&.912\\
\hline
RELE &.497&.905\\ \hline
SeG &.451&.913\\ \hline
$\text{BRE}$+ATT  &.457&\textbf{.917}\\
$\text{BRE}$+KA  &.480&\textbf{.917}\\
 \hline
$\text{BRE}$ &.625&.910\\
$\text{BRE}$+CE &\textbf{.630}&\textbf{.917}\\

\end{tabular}
\caption{AUC on NYT-FB60K and GIDS-FB8K.}
\label{nyt-gids}
\end{table}

\begin{figure}[htb]
\vspace{-0.8cm}
  \centerline{\includegraphics[width=0.48\textwidth]{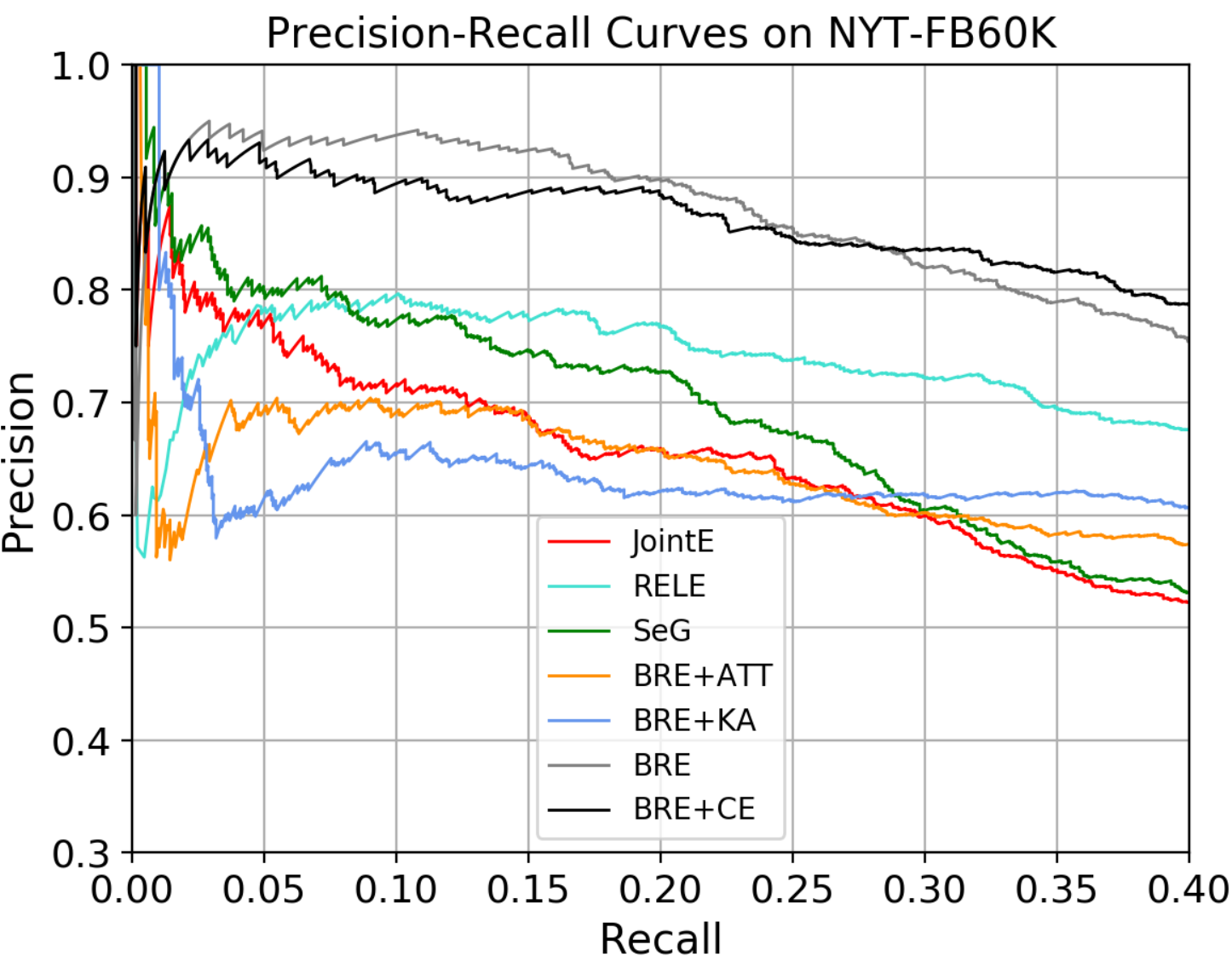}}

  \caption{Precision/recall curves on NYT-FB60K}
  \label{fig:prc}
 \vspace{-0.4cm}
\end{figure}

 Based on previous observations, we find that BRE and BRE+CE could avoid latent drawbacks of attention mechanism and have a stable performance on datasets with different noise patterns, thus they are competitive methods compared with prior baselines. To examine whether they work on the real-world Bag-level DSRE datasets, we compare our method to 3 previous baselines on NYT-FB60K \cite{han2018neural} and GIDS-FB8K \cite{jat2018improving}. We select JointE \cite{han2018neural}, RELE \cite{hu2019improving} and SeG \cite{li2020self} as baselines, because they achieve state-of-the-art performance on bag-level RE. To collect AUC results, we carefully re-run published codes of them using suggested hyperparameters from the original papers. We also draw precision-recall curves following prior works. As shown in Table \ref{nyt-gids} and Figure~\ref{fig:prc}, our method $\text{BRE}$+CE largely outperforms existing methods on NYT-FB60K and has comparable performance on GIDS-FB8K. Such result demonstrates that we avoid attention mechanism's latent drawback of hurting model's robustness. Furthermore, the model's improvement on NYT-FB60K is promising (around 13\% AUC). This is due to two reasons: (1) NYT-FB60K is a noisy dataset containing prevalent disturbing bags, which is similar to our synthesized datasets. (2)NYT-FB60K is highly imbalanced and most relation types only have limited training data, while all relation types in our balanced datasets have the same number of training examples; thus BRE+CE and BRE achieve much higher improvement on NYT-FB60K compared with synthesized datasets. In conclusion, the high performance not only validates our claim that attention module may not perform well on noisy and insufficient training data, but also verifies that our thorough analysis on attention and KG have practical significance.
 
\subsection{Effect of KG}
\begin{table}[htbp]
 \small
\centering
\begin{tabular}{c|p{0.9cm}|p{0.9cm}|p{0.9cm}}

\textbf{Model} & \textbf{BagRel}& \textbf{NYT}&\textbf{GIDS} \\ \hline
$\text{BRE}$+ATT &.878& .457&.917 \\
$\text{BRE}$+KA  &.932& .480&.917  \\
\hline
\hline
$\text{BRE}$ &.910& .625&.910  \\
$\text{BRE}$+CE &.915&.630&.917   \\
\end{tabular}
\caption{AUC test results of models on BagRel-Wiki73K, NYT-FB60K and GIDS-FB8K. In the BagRel column, all models are trained on $\text{train}_{\frac{1}{2},0}$.}  
\label{fewrel-3}
\vspace{-0.4cm}
\end{table}

From results in Table \ref{fewrel-3}, we provide a straight comparison between models with KG (BRE+KA, BRE+CE) and models without KG (BRE+ATT, BRE). Apparently, both methods of utilizing KG (combined with attention and concatenated as additional features) outperforms methods not using KG. This demonstrates the prior knowledge from KG is beneficial for relation extraction task. Except our naive BRE+CE, we expect that a carefully designed mechanism incorporating KG can lead to higher improvement. We leave it in the future.

\section{Conclusion}
In conclusion, we construct a set of datasets and propose a framework to quantitatively evaluate how attention module and KG work in the bag-level RE. Based on the findings, we demonstrate the effectiveness of a straightforward solution on this task. Experiment results well support our claims that the accuracy of attention mechanism depends on the noise pattern of the training set. In addition, although effectively selecting valid sentences, attention mechanism could harm model's robustness to noisy sentences and aggravate the data sparsity issue. As for KG's effects on attention, we observe that it promotes model's performance by enhancing its robustness with external entity information, instead of improving attention accuracy.

In the future, we are interested in developing a more general evaluation framework for other tasks, such as question answering, and improving the attention mechanism to be robust to noise and insufficient data, and an effective approach to incorporate the KG knowledge to guide the model training.

\section*{Acknowledgement}
This research/project is supported by NExT Research Centre. This research was also conducted in collaboration with SenseTime. This work is partially supported by A*STAR through the Industry Alignment Fund - Industry Collaboration Projects Grant, by NTU (NTU–ACE2020-01) and Ministry of Education (RG96/20), and by the National Research Foundation, Prime Minister’s Office, Singapore under its Energy Programme (EP Award No. NRF2017EWT-EP003-023) administrated by the Energy Market Authority of Singapore.
\bibliographystyle{acl_natbib}
\bibliography{acl2021}


\end{document}